\documentclass[a4paper, 10pt, conference]{ieeeconf}      % Use this line for a4 paper

\IEEEoverridecommandlockouts                              % This command is only needed if 
\usepackage{cite}
\usepackage[table]{xcolor}
\usepackage{times}
\usepackage{url}
\usepackage{latexsym}
\usepackage{booktabs}

\usepackage{todonotes}

\usepackage{amssymb}
\usepackage{amsmath}

\usepackage{cleveref}

\title{\LARGE \bf Detecting Agreement in Multi-party Conversational AI}

\author{\authorblockN{\bf Laura Schauer\authorrefmark{1}, Jason Sweeney\authorrefmark{1}, Charlie Lyttle\authorrefmark{1}, Zein Said\authorrefmark{1}, Áron Széles\authorrefmark{1},\\Cale Clark\authorrefmark{1}, Katie McAskill\authorrefmark{1}, Xander Wickham\authorrefmark{1}, Tom Byars\authorrefmark{1},\\ Daniel Hernandez Garcia\authorrefmark{2}, Nancie Gunson\authorrefmark{2}, Angus Addlesee\authorrefmark{2}, Oliver Lemon\authorrefmark{2}}
\authorblockA{\authorrefmark{1}School of Mathematical and Computer Sciences \\
Heriot-Watt University \\
{\tt {[laura.schauer212, jsnswny, clyttle0, zeinster2002, szelesaron39]}@gmail.com} \\
{\tt {[cc164, klm12, aw127, tjb10]}@hw.ac.uk}}
\authorblockA{\authorrefmark{2}The Interaction Lab \\
Heriot-Watt University\\
{\tt {[d.hernandez\_garcia, n.gunson, a.addlesee, o.lemon]@hw.ac.uk}}
}}

\begin{document}

\maketitle
\thispagestyle{empty}
\pagestyle{empty}

\begin{abstract}

Today, conversational systems are expected to handle conversations in multi-party settings, especially within Socially Assistive Robots (SARs). However, practical usability remains difficult as there are additional challenges to overcome, such as speaker recognition, addressee recognition, and complex turn-taking. In this paper, we present our work on a multi-party conversational system, which invites two users to play a trivia quiz game. The system detects users' agreement or disagreement on a final answer and responds accordingly. Our evaluation includes both performance and user assessment results, with a focus on detecting user agreement. Our annotated transcripts and the code for the proposed system have been released open-source on GitHub. 

\end{abstract}

%%%%%%%%%%%%%%%%%%%%%%%%%%%%%%%%%%%%%%%%%%%%%%%%%%%%%%%%%%%%%%%%%%%%%%%%%%%%%%%%
% INTRODUCTION PART
\section{Introduction}
\label{sec:introduction}

Socially assistive robots (SARs) are a crucial part of the future of many sectors, for example, education and healthcare \cite{gunson_visually_aware_2022}. SAR advancements are particularly important in the healthcare domain due to ageing populations. A serious lack of healthcare workers is already being experienced worldwide, with 10 million more health workers needed by 2030 \cite{cooper_ari_2020,Health_workforce_2023}. SARs can pose a solution to the problem by supporting healthcare in various ways, such as encouraging older people to keep living independently longer, or reducing caregiver burden \cite{cooper_ari_2020}.

Healthcare SAR scenarios frequently require these robots to handle multi-party interactions, as it is likely in these situations that a patient will be accompanied by family members or carers in a hospital waiting area \cite{Addlesee_Data_2023}. These groups may answer each other's questions, finish each other's sentences, and produce other multi-party specific phenomena \cite{addlesee2023multiparty}. While waiting for test results, patients often sit worrying with their companions \cite{huffman2023campfire}. In order to provide some light distraction, we propose a conversational system that plays a game of ``Who wants to be a millionaire?'' with two users. Our system impersonates the host while participants can collaborate on general knowledge trivia questions.

The conversational system is trained on multi-party human-human conversation data. We collected and annotated data from recordings of special ``Who wants to be a millionaire?'' episodes where two candidates collaborated to answer questions\footnote{\url{https://www.youtube.com/@Darren-wi2kt/videos}}. The system's aim is to guide users to collaborate and ultimately agree on an answer to the question. The code for our project is open-source and available on GitHub\footnote{\url{https://github.com/JsnSwny/tarrant}}.

Our system is developed to be ported onto an ARI robot \cite{cooper_ari_2020} in the course of the SPRING project\footnote{\url{https://spring-h2020.eu/}}. This project encompasses several European universities collaborating on evolving socially assistive robots in public environments. All the advancements are integrated and deployed on an ARI robot in a hospital memory clinic in Paris with real patients.

% LIT REVIEW PART 
\section{Background}
\label{sec:background}
%CUT BY CHARLIE
% \subsection{Socially Assistive Robots}
% \label{subsec:socially_assistive_robots}
% For healthcare, as well as for any other sector, the difficulty of successfully designing SARs lies in creating robots that can effectively converse with humans and adhere to social norms \cite{moujahid_multi_party_2022}. The more expressive a robot is, the more it will be perceived as intelligent, conscious and polite \cite{moujahid_multi_party_2022}.

% The SPRING project conducts research on a SAR robot deployed in an eldercare hospital reception area \cite{addlesee_comprehensive_2020}. The conversational system is deployed on the humanoid ARI robot produced by Pal Robotics \cite{palrobot}. ARIs capabilities can be extended with custom AI algorithms, in the case of SPRING-ARI a visual perception system, a dialogue system, and a social interaction planner \cite{addlesee_comprehensive_2020}. While the SPRING-ARI system successfully demonstrates that task-based, social and visually grounded dialogue can be combined with physical actions, it still lacks the ability to handle conversations with more than one person simultaneously \cite{addlesee_comprehensive_2020}.

\subsection{Multi-party conversational systems}
\label{subsec:multi_party}
In general, most conversational systems are only able to handle dyadic (two-party) conversations \cite{Mahajan_Santhanam_Shaikh_2022,moujahid_multi_party_2022}. The endeavour to create naturally interactive conversational systems becomes considerably more difficult when dealing with multi-party interactions \cite{Addlesee_Data_2023}. Compared to handling dyadic interactions, handling multi-party conversations includes more complex challenges such as: coordination of turn-taking, speaker recognition, addressee recognition, and complex response generation depending on whom the system is addressing (with these latter three tasks being commonly referred to collectively as ``who says what to whom?'') \cite{Addlesee_Data_2023,Johansson_Skantze_2015, traum2004issues, gu2021mpc}. 

Turn-taking poses a central problem. In dyadic conversations, there are only two roles a participant can take: speaker or listener, hence it is clear when and to whom the turn is yielded. In multi-party conversations, participants can take multiple roles, and address individuals or groups, so turn-taking needs to be coordinated \cite{Johansson_Skantze_2015}.

Most existing multi-party conversational research has focused on the sub-tasks of ``Who says what to whom''. Mahajan et al, identified 15 papers in the last decade focusing on one or more of these sub-tasks within an English-speaking context \cite{Mahajan_Santhanam_Shaikh_2022}. Centralised evaluation criteria are relatively new, and most existing systems are not consistently evaluated. One previous study, conducted by Skantze et al, required users to collaborate with a robot in a multi-party setting. The users' tasks were to play a card game with the Furhat robot, where they had to sort the cards in a specific order. However, the research focused on turn-taking cues rather than the system's performance in terms of aiding users to collaborate, or detecting agreement between users \cite{Skantze_Johansson_Beskow_2015,Furhat_2023}.

% DATA COLLECTION
\section{Data Collection}
\label{sec:data_collection}

An important point to consider in multi-party training data is that most are human-human. This can pose a problem, as systems trained on human-human conversations may generate realistic looking conversations, but do not inherently have an incentive to help users reach their goals \cite{Addlesee_Data_2023}. However, in our scenario, the system has the clear goal of guiding the participants to find an answer to a question. To our knowledge, there are no multi-party corpora available for this use case, therefore data collection was performed by us. We first collected 115 available recordings of ``Who Wants to Be a Millionaire'' with two participants, which were transcribed using the YouTube API\footnote{\url{https://pypi.org/project/youtube-transcript-api/}}. To annotate these transcripts, we used the set of annotations shown in \Cref{table:intents}. This list allowed us to capture as much information as possible, without saturating the data. The annotated transcripts are publicly available on GitHub\footnote{\url{https://github.com/lauraschauer/tar-transcripts}}. 

\noindent
\begin{table}
  \label{table:intents}
  \begin{tabular}{@{} l p{0.65\columnwidth} @{}}
    \toprule
    \textbf{Intent}      & \textbf{Description}                                                     \\
    \midrule
    \textit{Host (System) Intents}   &                                       \\[3pt]
    question             & The system presents the question                                         \\
    options              & The system presents the options                                          \\
    confirm-agreement    & The system tries to confirm the final answer with participants           \\
    accept-answer        & System considers answer the final answer                                 \\
    \midrule
    \textit{User Intents}       &                                            \\[3pt]
    chit-chat            & Speech not related to the quiz                                           \\
    offer-answer()       & A player presents an answer to the other player                          \\
    offer-to-answer      & A  player signals that they know the answer                              \\
    agreement            & Agreement between players about the answer                               \\
    ask-agreement        & A player asks the other player for confirmation on their proposed answer \\
    final-answer()       & Players give final answer                                                \\
    confirm-final-answer & Participants confirm their answer is final                               \\
    \bottomrule
  \end{tabular}
  \caption{Intents used for Data Annotation}
\end{table}

\begin{figure}
  \centering
  \includegraphics[width=0.95\columnwidth]{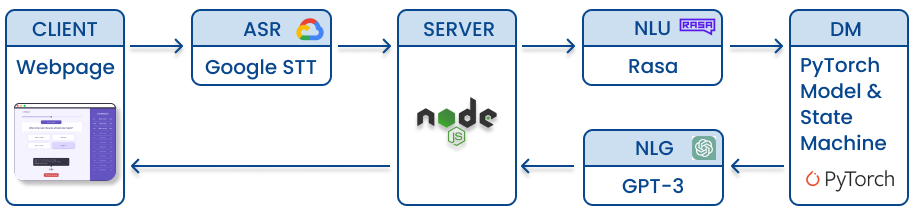}
  \caption{Architecture of the conversational system}
  \label{fig:system_architecture}
\end{figure}

%\subsection{Cohen's Kappa Coefficient}
%To ensure reliability and consistency, Cohen's kappa was calculated for a sample of the completed transcripts. It measures the reliability between raters on categorical data, while accounting for agreement happening by chance \cite{Cohen_1960}. 15\% of the total transcripts were re-annotated by others, resulting in a Kappa score of 0.9601. This score is interpreted as ``almost perfect agreement'' \cite{McHugh_2012}, and the annotation of transcripts can be concluded as reliable.

% \begin{equation}
%   \begin{split}
%     KappaScore & = \frac{Agree - ChanceAgree}{1 - ChanceAgree} \\
%     KappaScore & = \underline{\underline{0.9601}}
%   \end{split}
% \end{equation}

% \noindent
% A Kappa score of 0.9601 is interpreted as ``almost perfect agreement'' \cite{McHugh_2012}. From this, the annotation of transcripts can be concluded as reliable.

% DESIGN & IMPLEMENTATION
\section{Design and Implementation}
\label{sec:implementation}

Our conversational system consists of several parts, the modular architecture is shown in \Cref{fig:system_architecture}. This section outlines each unit and describes how our system manages the process from the users' utterances to selecting its own response.

% AUTOMATIC SPEECH RECOGNITION PART
\subsection{Automatic Speech Recognition}
\label{subsec:asr}

In order to reliably record both parties within the conversation, we bypass diarisation \cite{addlesee_comprehensive_2020, addlesee2023detecting} by using two separate microphones. Two users connect to the web app on separate laptops, and their integrated microphones record the conversation.

A drawback of this set-up is that the microphones can pick up both users' voices, as we observed in the system's diarisation transcripts. This was mitigated by moving the microphones further apart, and calibrating them. Nevertheless, the issue still arose during our evaluation, results of which are discussed in \Cref{sec:evaluation}.

The real-time transcription of the users' utterances are then passed onto the natural language understanding (NLU) and Dialogue Manager (DM).

\subsection{Dialogue Management}
\label{subsec:dialogue_management}
The dialogue manager is responsible for selecting an adequate response at any given moment. It is also tasked with asking questions which it obtains from the OpenTrivia Database, a user-contributed trivia question database under the Creative Commons Licence \cite{Open_trivia_2023}. The dialogue manager is composed of two components: a state machine and a Pytorch model. This approach was chosen because, despite involving a conversation which requires intelligent behaviour from the dialogue manager, the game itself is defined by a set of rigid rules:

\begin{itemize}
  \setlength\itemsep{0.05em}
  \item One round is comprised of ten questions.
  \item The system must drive the conversation by passing from one question to the next.
  % CUT BY LAURA
  % \item The game must end after a specified time.
\end{itemize}

\noindent The hard-coded logic of the state machine occasionally overwrites the model's output, ensuring the game rules are followed, and guaranteeing the movement from state to state.

% NEURAL NETWORK PART
\subsubsection{Neural Network}
\label{subsec:nn}
The LSTM neural network model is designed to manage dialogues of three-person conversations. To do this, it has two inputs, the intent from the NLU and the user ID (user 1, user 2, or host). These inputs were both one-hot encoded, the intent being a vector of size 14 (one for each intent) and the user number being a vector of size 3. These two vectors were concatenated to produce an input vector of size 17. The network was then trained to predict the host's response for a given intent and user in a sequence. This output is a vector of size 5, one for each of the 4 system intents plus one for \verb|no response| (see \Cref{table:intents}).

The model was trained using the dataset, which we had originally labelled for NLU purposes. This provided human labelled sequences of intents and user IDs for each question. Each question would have its intents and user IDs input into the network sequentially, with the model's memory being cleared after every question. Every output from the system which is not \verb|no response| would be input back into the network, allowing for the system to speak multiple lines of dialogue in succession. At the start of each question, the system is set up by first inputting the host having given the \verb|question| intent, to which it will always output the \verb|options| intent.
%This is required since every question in the training data is started by the host asking the question, then giving the options for the question.

\begin{figure}
\centering
  \includegraphics[width=0.7\columnwidth]{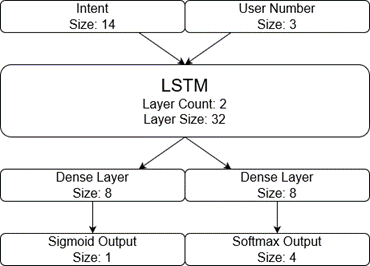}
  \caption{Architecture of PyTorch Model}
  \label{fig:model_architecture}
\end{figure}

The primary difficulty of training was the imbalance in the dataset. Since most of the time, the host will be only listening, \verb|no response| is by far the most common output. This means that having the output as a single SoftMax vector of size 5 would be difficult, as the model would quickly learn to always output \verb|no response|. Traditional undersampling and oversampling methods would also be difficult to make use of, since the \verb|no response| output exists as part of a real conversation, which must be fed into the network sequentially. Similarly, uncommon system responses could not be oversampled, as they exist as part of a real question. Artificially increasing their frequency would alter the questions, diminishing the integrity of the data. To overcome this, a separate Sigmoid output was used for \verb|no response|, as shown in \Cref{fig:model_architecture}. To train this model, whenever \verb|no response| from the host was expected, the loss from the SoftMax vector was set to zero, otherwise L2 loss was used for all outputs. The system was implemented in PyTorch, with the model being trained for 30 epochs with a learning rate of $5 \times 10^{-4}$.

In future, this architecture could be used for handling multi-person dialogues with more intents or more users by simply scaling the inputs, outputs, and hidden layers accordingly. The model could also be altered to allow for more inputs, such as pauses or other non-verbal cues.

% STATE MACHINE PART
\subsubsection{State Machine}
\label{subsec:dm}

\begin{figure}
    \centering
    \includegraphics[width=0.95\columnwidth]{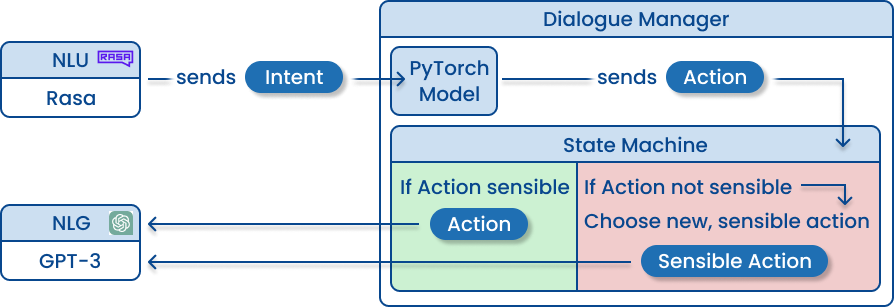}
    \caption{State Machine Behaviour}
    \label{fig:dm_interaction}
\end{figure}

An intent received from the NLU unit is first input into the PyTorch model, which outputs an action (\Cref{fig:dm_interaction}). Afterwards, the state machine checks if the output of the PyTorch model represents a sensible action. In case there are logical errors, the output will be overridden. For example, if the NLU detects that the user intends to confirm their final answer, but no answer has yet been given, the PyTorch model may nonetheless decide on \verb|accept-answer|. As no answer has been given yet, the state machine will overwrite this output.

Another reason for using a state machine was the limited amount of transcription data. The PyTorch model learned from the transcript data, therefore it depended on a sufficient amount of transcript data to function adequately. The state machine extends the set of actions by ten additional intents, which are listed in \Cref{table:state_machine_intents}. This allowed us to program explicitly how the system responds to user intents, thus eliminating the need for more transcript data.

% CHANGED BY LAURA
% \begin{figure}
%   \centering
%   \includegraphics[width=0.9\columnwidth]{images/jacks_figure.png}
%   \caption{State Machine behaviour}
%   \label{fig:jacks_figure}
% \end{figure}

% CUT BY LAURA
% The state machine's configuration is defined inside a JSON file, as shown in \Cref{fig:dm_interaction}. The figure shows the behaviour of the state machine when the state is \verb|question|. An example of previously recorded values influencing the flow can be seen inside \verb|agreement|. When the action decided by the chatbot is overridden by the state machine, it is done according to this configuration. The action chosen by the state machine is a string of JavaScript code, and is found in the configuration object \verb|flow| at \verb|flow[state][intent]|, where \verb|state| is the current state of the machine and \verb|intent| is the name of the most recent user intent.

Throughout the course of a game, the dialogue manager stores relevant values, such as answers offered by the user. These values influence the behaviour of the state machine. An example of this is the \verb|seek-confirmation| state when the host asks if the users would like to lock in an answer suggested by them. If a user declines, the host checks if the rejected answer is the same as the one just offered by the user. If it matches, then the host assumes that the user does not want to lock in that answer. However, if the answer rejected by the user is another one, then the host will assume that they do indeed wish to proceed with their offered answer.

\noindent
\begin{table}
  \label{table:state_machine_intents}
  \begin{tabular}{@{} l p{0.6\columnwidth} @{} }
    \toprule
    \textbf{Intent}           & \textbf{Description}      \\
    \midrule
    acknowledge-reject-option & The system acknowledges that a player has dismissed an option and may ask them for a reason.                                              \\

    end-of-game               & The system announces the end of the game and informs the players of how they did.                                                         \\

    offer-generic-guidance    & The system offers advice or encouragement to the players.                                                                                 \\

    question-brief            & The system restates the current question number and question prize.                                                                       \\

    repeat-answer             & The system repeats a player's answer back to them.                                                                                        \\

    return-to-question        & The system returns to the question state, becoming less insistent on locking in answers suggested.                                        \\

    say-correct               & The system states that the player's final answer was correct.                                                                            \\

    say-incorrect             & The system states that the player's final answer was incorrect.                                                                          \\

    seek-confirmation         & The system asks the players if they are wishing to lock in a suggested answer as final.                                                   \\

    seek-direct-answer        & The system asks the players to state their answer. This is done by the system when it suspects that it may have missed an answer offered. \\
    \bottomrule
  \end{tabular}
  \caption{Additional State Machine Intents}
\end{table}
A more secure approach to this situation would involve taking a record of all the answers ruled out by the users. The host would then lock in a final answer after a rejected answer if and only if the other two possible answers had also been previously rejected. We will implement this in a future version, as it would reduce the risk of the host accepting answers while the users are still deciding.

Once the DM has identified an appropriate response, the intent is passed to the Natural Language Generation (NLG).
%CUT BY CHARLIE
\subsection{Natural Language Generation}
\label{subsec:nlg}

In order for natural and realistic responses to be generated by the Natural Language Generation (NLG) we used OpenAI's API, specifically “gpt-3.5-turbo”, in order to generate 50 unique options for each possible response from the game `host'. Each generated response was checked manually for hallucinations, offence, and accuracy.

%CUT BY CHARLIE
%To make the system feel more unique and less robotic, we made use of Natural Language Generation (NLG) for the phrases that the “host” says to the participants. We used OpenAI's API, specifically “gpt-3.5-turbo”, the same version that is used in ChatGPT. This allowed us to prompt for different outputs for the system that convey the same information. For example, when receiving a correct answer, “Yes, that's it! Well done!” and “You got it! Great job!” are both possible outputs. There are 50 different options for each response the “host” can say.

% To further improve on NLG within this system, some content moderation could be performed on the generations, to ensure that there are no inappropriate outputs, this can be done entirely within OpenAI's API.  Also, the system could be updated to make use of GPT-4, as GPT-4 works more effectively to avoid inappropriate content and has greater problem solving abilities, however, at this current time it is not publicly available \cite{GPT_4_2023,OpenAI_2023}.

% EVALUATION PART
\section{Evaluation}
\label{sec:evaluation}
\subsection{Experiment Layout}
\label{subsec:experiment_layout}
We evaluated both subjective and objective measures of the system's performance. The subjective measures included the user's enjoyment and perception of the system's natural behaviour, while the objective measures focused on the agreement detection accuracy. The agreement detection accuracy indicates the frequency of the system correctly detecting users' intents to submit a final answer.

\begin{figure}
  \centering
  \includegraphics[width=0.8\columnwidth]{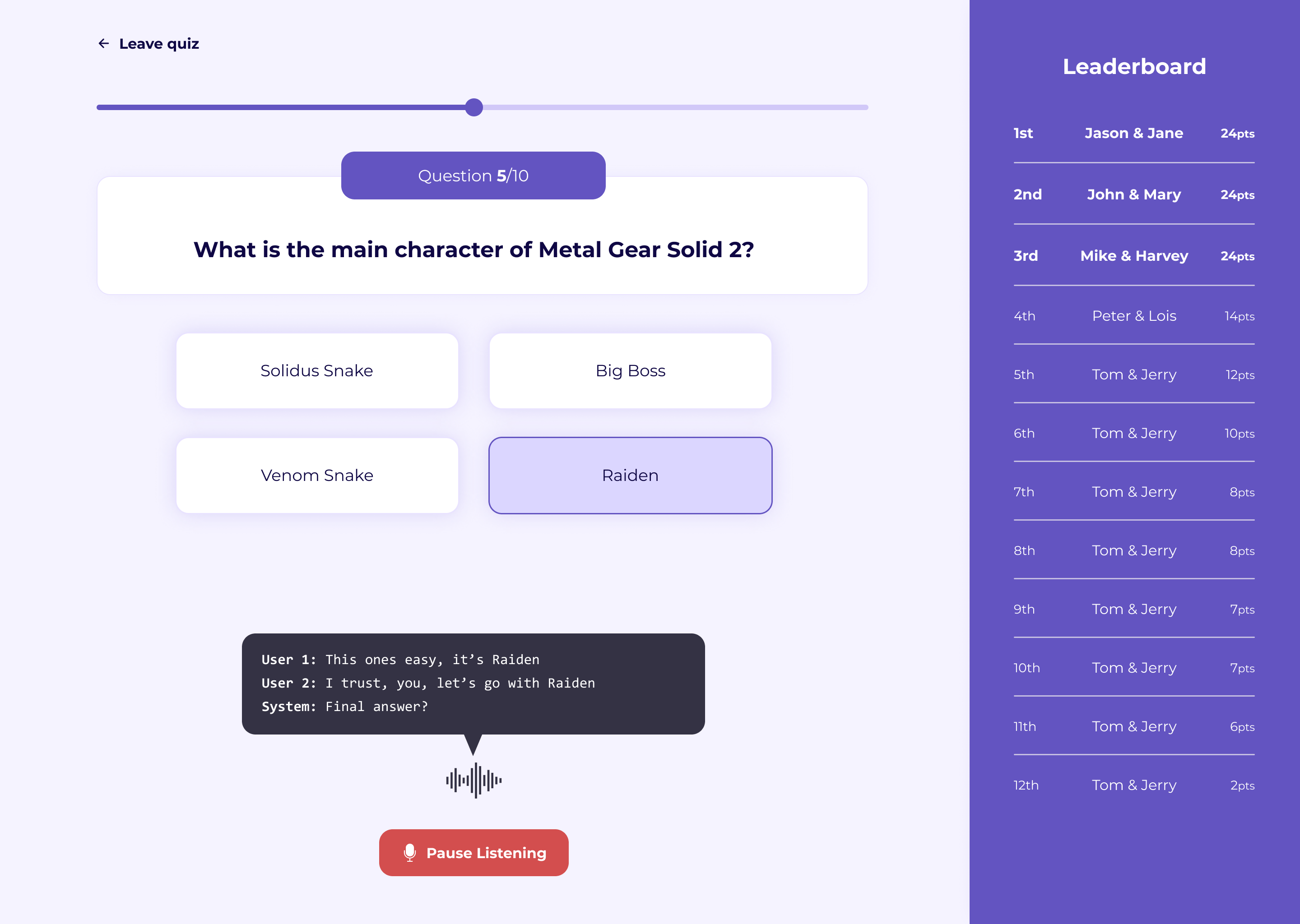}
  \caption{Main Webpage}
  \label{fig:webpage}
\end{figure}

% The evaluation took place in two iterations in a between-subjects design, both iterations with three pairs of participants. 
We evaluated our system with six pairs of participants. Each participant sat in front of a laptop, where they could see the web page with the question and answer options on the screen (see \Cref{fig:webpage}). 
Each pair of participants played one round of the quiz, answering a maximum of 10 questions. Afterwards, they were asked to complete a questionnaire about their experience. They rated the system using a five-point Likert scale.

\subsection{Results}
\label{subsec:results}

\subsubsection{User Satisfaction}
The system was well received as 100\% of users were satisfied with it (\(score \geq 3\)). Regarding the system's behaviour, 50\% of participants indicated that the system appeared natural. Furthermore, 83\% of participants reported enjoying the experience. The likert responses returned a mean value of 3.81 displaying a generally positive reaction to the system. As well as this, the likert responses gave a standard deviation value of 1.03 showing a consistent level of satisfaction with the system between users.

%CUT BY CHARLIE
%\subsubsection{Fuzzy String Matching}
%\label{subsec:fuzzy_string}
%Fuzzy string matching is a computational technique to approximate string matches. It does so by tokenizing both strings and comparing the similarity of common tokens of both strings. In our system, we employed fuzzy string matching to identify potential answers based on the participants' speech. The selection of an optimal similarity threshold was critical for performance. We performed a within-subject analysis, comparing a 0.5 and 0.7 threshold. The 0.5 threshold yielded the best performance, as all users' answers were matched with an answer option. With a 0.7 threshold, 33\% of users' answers were not matched due to imperfect transcriptions. A 0.5 threshold is more effective at mitigating transcription errors compared to a 0.7 threshold.

\subsubsection{Agreement Detection}
\begin{figure}
  \centering
  \includegraphics[width=0.8\columnwidth]{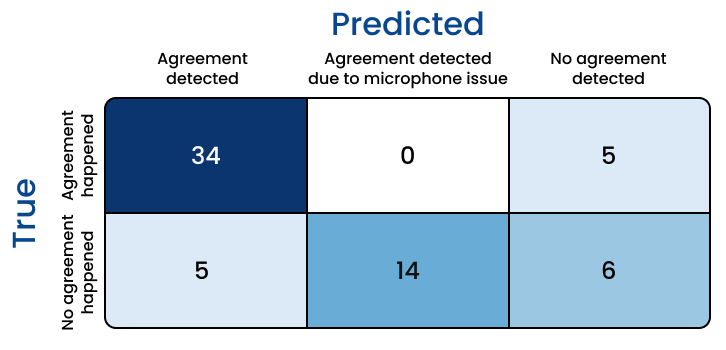}
  \caption{Agreement Detection Confusion Matrix}
  \label{fig:confusion-matrix}
\end{figure}

The confusion matrix presents the real agreements against the system's agreement predictions across all 6 evaluations (\Cref{fig:confusion-matrix}). We deduced these values manually from recordings of each evaluation and the system's speech transcription. The True Positive value is when the system accurately identified agreement, the False Positive value is when it identified agreement that has not happened with a separate column for the microphone issue. A False Negative is when the system missed an agreement, and the True Negative value is when users did not wish to lock in an answer and the system recognised that.

The main issue with our system are false positive predictions. These are caused by both laptop microphones of the participants picking up the same sentence, though only one participant had spoken. The system would then class this as agreement. False negatives occurred when the answer terms were difficult to pronounce and the system struggled to pick them up. By adding the ``Agreement detected due to microphone issue'' column, mistakes caused by the microphone issue are factored out and lead to an agreement detection accuracy of 80\%. The precision, recall, and f1-score were all 0.87, excluding the errors due to the microphone issue mentioned above.
% and an error rate of 20\%. 

\section{Conclusion and Future Work}
\label{sec:conclusion}

In this paper, we have presented a conversational system to play a game of ``Who wants to be a millionaire?'' in a multi-party setting. The system is able to detect when users agree on a final answer, or will continue asking for an answer to encourage users to reach a conclusion. The question and answer options are displayed on a web application, which mimics the display in the ARI robot's screen.
To evaluate the system, we conducted an experiment assessing users' perception of the system, as well as its performance. If the microphone issue can be avoided, using social robots designed for speech interaction \cite{addlesee2023building}, the system reaches an agreement detection accuracy of 80\% and has a high rate of user satisfaction.

In order to improve the system and test its effectiveness further, we are implementing the system on the ARI social robot \cite{cooper_ari_2020}, and will test it with human participants. Our code was designed to enable seamless porting to the robot, but we will examine the acoustics of its microphones and speakers in the test environment to avoid interaction failure \cite{addlesee2023building}.

\bibliographystyle{ieeetran}
\bibliography{references}

%%%%%%%%%%%%%%%%%%%%%%%%%%%%%%%%%%%%%%%%%%%%%%%%%%%%%%%%%%%%%%%%%%%%%%%%%%%%%%%%
%\section*{APPENDIX}

%\subsection{NLU Intent Recognition Confusion Matrix}
%\label{app:cm_intent_recognition}

%\begin{figure}[h!]
 % \centering
  %\includegraphics[width=0.8\columnwidth]{images/intent_confusion_matrix.jpg}
  %\caption{Confusion Matrix of NLU intent recognition}
%\end{figure}

%\newpage

%\subsection{Screenshot of System's Webpage}
%\label{app:system_webpage}

%\begin{figure}[h!]
%  \centering
  %\includegraphics[width=0.8\columnwidth]{images/webpage.png}
  %\caption{Screenshot of System's Webpage}
%\end{figure}

% \section*{ACKNOWLEDGMENT}

% The preferred spelling of the word acknowledgment in America is without an e after the g. Avoid the stilted expression, One of us (R. B. G.) thanks . . .  Instead, try R. B. G. thanks. Put sponsor acknowledgments in the unnumbered footnote on the first page.

%%%%%%%%%%%%%%%%%%%%%%%%%%%%%%%%%%%%%%%%%%%%%%%%%%%%%%%%%%%%%%%%%%%%%%%%%%%%%%%%

\end{document}